# Acquiring Correct Knowledge
# for Natural Language Generation

**Ehud Reiter**                                        EREITER@CSD.ABDN.AC.UK
**Somayajulu G. Sripada**                              SSRIPADA@CSD.ABDN.AC.UK
*Department of Computing Science,*
*University of Aberdeen, Aberdeen AB24 3UE, UK*

**Roma Robertson**                                     ROMA.ROBERTSON@ED.AC.UK
*Division of Community Health Sciences - General Practice Section*
*University of Edinburgh*
*Edinburgh EH8 9DX, UK*

## Abstract

Natural language generation (NLG) systems are computer software systems that produce texts in English and other human languages, often from non-linguistic input data. NLG systems, like most AI systems, need substantial amounts of knowledge. However, our experience in two NLG projects suggests that it is difficult to acquire correct knowledge for NLG systems; indeed, every knowledge acquisition (KA) technique we tried had significant problems. In general terms, these problems were due to the complexity, novelty, and poorly understood nature of the tasks our systems attempted, and were worsened by the fact that people write so differently. This meant in particular that corpus-based KA approaches suffered because it was impossible to assemble a sizable corpus of high-quality consistent manually written texts in our domains; and structured expert-oriented KA techniques suffered because experts disagreed and because we could not get enough information about special and unusual cases to build robust systems. We believe that such problems are likely to affect many other NLG systems as well. In the long term, we hope that new KA techniques may emerge to help NLG system builders. In the shorter term, we believe that understanding how individual KA techniques can fail, and using a mixture of different KA techniques with different strengths and weaknesses, can help developers acquire NLG knowledge that is mostly correct.

## 1. Introduction

Natural language generation (NLG) systems use artificial intelligence (AI) and natural language processing techniques to automatically generate texts in English and other human languages, typically from some non-linguistic input data (Reiter & Dale, 2000). As with most AI systems, an essential part of building an NLG system is knowledge acquisition (KA), that is acquiring relevant knowledge about the domain, the users, the language used in the texts, and so forth.

KA for NLG can be based on structured expert-oriented techniques, such as think-aloud protocols and sorting, or on machine learning and corpus analysis, which are currently very popular in other areas of Natural Language Processing. We have used both types of techniques in two NLG projects that included significant KA efforts – STOP (Reiter, Robertson, & Osman, 2003), which generated tailored smoking cessation letters, and SUMTIME-MOUSAM (Sripada, Reiter, Hunter, Yu, & Davy, 2001), which generated weather forecasts. In both projects, and for all techniques tried, the main problem turned out to be knowledge quality; evaluation and validation exercises identified flaws in the knowledge acquired using every technique. The flaws were due to a variety of factors, but perhaps the basic underlying reason for them was the nature of the writing tasks we were attempting to automate. They were:

- complex (as are many tasks that involve interacting with humans): hence a lot of knowledge was needed to cover the numerous special cases and unusual circumstances;

- sometimes novel (not done by humans): hence sometimes there were no experts at the task as a whole, and no existing corpora of texts to analyse;

- poorly understood: hence we did not have good theoretical models to structure the knowledge being acquired, and fill in gaps in the knowledge acquired from experts or corpora; and

- ambiguous (allowed multiple solutions): hence different experts and corpus authors produced very different texts (solutions) from the same input data.

These problems of course occur to some degree in KA for other expert system and natural language processing tasks, but we believe they may be especially severe for NLG.

We do not have a good solution for these problems, and indeed believe that KA is one of the biggest problems in applied NLG. After all, there is no point in using AI techniques to build a text-generation system if we cannot acquire the knowledge needed by the AI techniques.

In the longer term, more basic research into KA for NLG is badly needed. In the shorter term, however, we believe that developers are more likely to acquire correct knowledge when building an NLG system if they understand likely types of errors in the knowledge acquired from different KA techniques. Also, to some degree the different KA techniques we have tried have complementary strengths and weaknesses; this suggests using a variety of different techniques, so that the weaknesses of one technique are compensated for by the strengths of other techniques.

In the remainder of this paper we give background information on NLG, KA, and our systems; describe the various KA techniques we used to build our systems and the problems we encountered; and then discuss more generally why KA for NLG is difficult and how different KA techniques can be combined.

## 2. Background

In this section we give some background information on natural language generation and knowledge acquistion and validation. We also introduce and briefly describe the STOP and SumTime-Mousam systems.

## 2.1 Natural Language Generation

Natural Language Generation is the subfield of artificial intelligence that is concerned with automatically generating written texts in human languages, often from non-linguistic input data. NLG systems often have three stages (Reiter & Dale, 2000):

- *Document Planning* decides on the content and structure of the generated text; for example that a smoking-cessation letter should start with a section that discusses the pros and cons of smoking.

- *Microplanning* decides on how information and structure should be expressed linguistically; for example, that the phrase *by mid afternoon* should be used in a weather report to refer to the time 1500.

- *Surface Realisation* generates an actual text according to the decisions made in previous stages, ensuring that the text conforms to the grammar of the target language (English in our systems).

NLG systems require many types of knowledge in order to carry out these tasks. In particular, Kittredge, Korelsky, and Rambow (1991) point out that NLG systems need domain knowledge (similar to that needed by expert systems), communication knowledge (similar to that needed by other Natural Language Processing systems), and also domain communication knowledge (DCK). DCK is knowledge about how information in a domain is usually communicated, including standard document structures, sublanguage grammars, and specialised lexicons. DCK plays a role in all aspects of language technology (for example, a speech recogniser will work better in a given domain if it is trained on a corpus of texts from that domain), but it may be especially important in NLG.

## 2.2 Knowledge Acquisition and Validation

Knowledge acquisition is the subfield of artificial intelligence that is concerned with acquiring the knowledge needed to build AI systems. Broadly speaking the two most common types of KA techniques are:

- Techniques based on working with experts in a structured fashion, such as structured interviews, think-aloud protocols, sorting, and laddered grids (Scott, Clayton, & Gibson, 1991; Buchanan & Wilkins, 1993); and

- Techniques based on learning from data sets of correct solutions (such as text corpora); these are currently very popular in natural language processing and used for many different types of knowledge, ranging from grammar rules to discourse models (for an overview, see Jurafsky & Martin, 2000).

There are of course other possible KA techniques as well, including directly asking experts for knowledge, and conducting scientific experiments. Some research has been done on evaluating and comparing KA techniques, but such research can be difficult to interpret because of methodological problems (Shadbolt, O'Hara, & Crow, 1999).

Research has also been done on verifying and validating knowledge to check that it is correct (Adelman & Riedel, 1997). Verification techniques focus on detecting logical anomalies and inconsistencies that often reflect mistakes in the elicitation or coding process; we will not further discuss these, as such errors are not our primary concern in this paper. Validation techniques focus on detecting whether the knowledge acquired is indeed correct and will enable the construction of a good system; these are very relevant to efforts to detect problems in knowledge acquired for NLG. Adelman and Riedel (1997) describe two general types of validation techniques: (1) having experts check the acquired knowledge and built systems, and (2) using a library of test cases with known inputs and outputs. In other words, just as knowledge can be acquired from experts or from data sets of correct solutions, knowledge can also be validated by experts or by data sets of correct solutions. Knowledge can also be validated experimentally, by determining if the system as a whole works and has the intended effect on its users. Of course care must be taken that the validation process uses different resources than the acquisition process. For example, knowledge acquired from an expert should not be validated by that expert, and knowledge learned from a data set should not be validated by that data set.

There has not been a great deal of previous research on knowledge acquisition for NLG; Reiter, Robertson, and Osman (2000) summarise previous efforts in this area. Generally corpus analysis (analysis of collections of manually written texts) has been the most popular KA technique for NLG, as in other areas of Natural Language Processing, although sometimes it is supplemented by expert-oriented techniques (Goldberg, Driedger, & Kittredge, 1994; McKeown, Kukich, & Shaw, 1994). Walker, Rambow, and Rogati (2002) have attempted to learn NLG rules from user ratings of generated texts, which can perhaps be considered a type of experiment-based KA.

## 2.3 STOP

STOP (Reiter, Robertson, & Osman, 2003) is an NLG system that generates tailored smoking-cessation letters. Tailoring is based on a 4-page multiple-choice questionnaire about the smoker's habits, health, concerns, and so forth. An extract from a questionnaire is shown in Figure 1, and an extract from the STOP letter generated from this questionnaire is shown in Figure 2 (we have changed the name of the smoker to preserve confidentiality). From a KA perspective, the most important knowledge needed in STOP is what content and phrasing is appropriate for an individual smoker; for example,

- What information should be given in a letter? The example letter in Figure 2, for instance, emphasises things the smoker dislikes about smoking, confidence building, and dealing with stress and weight gain; but it does not recommend specific techniques for stopping smoking.

- Should a letter adopt a positive 'you'll feel better if you stop' tone (as done in the letter in Figure 2), or should it adopt a negative 'smoking is killing you' tone?

STOP was never operationally deployed, but it was tested with real smokers in a clinical trial, during which 857 smokers received STOP letters (Lennox, Osman, Reiter, Robertson, Friend, McCann, Skatun, & Donnan, 2001). This evaluation, incidentally, showed that STOP letters were no more effective than control non-tailored letters.

## SMOKING QUESTIONNAIRE

Please answer by marking the most appropriate box for each question like this: ☒

---

**Q1 Have you smoked a cigarette in the last week, even a puff?**

YES ☒                          NO ☐

Please complete the following questions          Please return the questionnaire unanswered in the envelope provided. Thank you.

---

**Please read the questions carefully.**     If you are not sure how to answer, just give the best answer you can.

---

Q2   **Home situation**:

Live alone ☒          Live with husband/wife/partner ☐          Live with other adults ☐          Live with children ☐

Q3   **Number of children** under 16 living at home     ………**0**……… boys     ………**0**……. girls

Q4   **Does anyone else in your household smoke?**   *(If so, please mark all boxes which apply)*

husband/wife/partner ☐          other family member ☐          others ☐

---

Q5   **How long have you smoked for?**   …**20**… years

Tick here if you have smoked for less than a year     ☐

---

Q6   **How many cigarettes do you smoke in a day?**   *(Please mark the amount below)*

Less than 5 ☐     5 – 10 ☐     11 – 15 ☒     16 – 20 ☐     21 - 30 ☐     31 or more ☐

Q7   **How soon after you wake up do you smoke your first cigarette?** *(Please mark the time below)*

Within 5 minutes ☐     6 - 30 minutes ☒     31 - 60 minutes ☐     After 60 minutes ☐

Q8   **Do you find it difficult not to smoke in places where it is forbidden   eg in church, at the library, in the cinema?**          YES ☒   NO ☐

Q9   **Which cigarette would you hate most to give up?**          The first one in the morning ☒
Any of the others ☐

Q10   **Do you smoke more frequently during the first hours after waking than during the rest of the day?**          YES ☐   NO ☒

Q11   **Do you smoke if you are so ill that you are in bed most of the day?**          YES ☐   NO ☒

---

Q12

**Are you intending to stop smoking in the next 6 months?**          YES ☐

NO ☒

Q13   **If yes, are you intending to stop smoking within the next month?**

YES ☐     NO ☐

Q14   **If no, would you like to stop smoking if it was easy?**

YES ☐     Not Sure ☒     NO ☐

Figure 1: First page of example smoker questionnaire

## Smoking Information for Heather Stewart

### You have good reasons to stop...

People stop smoking when they really want to stop. It is encouraging that you have many good reasons for stopping. The scales show the good and bad things about smoking for you. They are tipped in your favour.

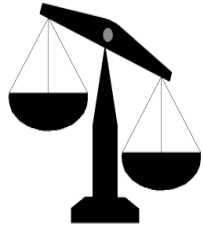

THINGS YOU LIKE

it's relaxing
it stops stress
you enjoy it
it relieves boredom
it stops weight gain
it stops you craving

THINGS YOU DISLIKE

it makes you less fit
it's a bad example for kids
you're addicted
it's unpleasant for others
other people disapprove
it's a smelly habit
it's bad for you
it's expensive
it's bad for others' health

### You could do it...

Most people who really want to stop eventually succeed. In fact, 10 million people in Britain have stopped smoking - and stayed stopped - in the last 15 years. Many of them found it much easier than they expected.

Although you don't feel confident that you would be able to stop if you were to try, you have several things in your favour.

- You have stopped before for more than a month.
- You have good reasons for stopping smoking.
- You expect support from your family, your friends, and your workmates.

We know that all of these make it more likely that you will be able to stop. Most people who stop smoking for good have more than one attempt.

### Overcoming your barriers to stopping...

You said in your questionnaire that you might find it difficult to stop because smoking helps you cope with *stress*. Many people think that cigarettes help them cope with stress. However, taking a cigarette only makes you feel better for a short while. Most ex-smokers feel calmer and more in control than they did when they were smoking. There are some ideas about coping with stress on the back page of this leaflet.

You also said that you might find it difficult to stop because you would *put on weight*. A few people do put on some weight. If you did stop smoking, your appetite would improve and you would taste your food much better. Because of this it would be wise to plan in advance so that you're not reaching for the biscuit tin all the time. Remember that putting on weight is an overeating problem, not a no-smoking one. You can tackle it later with diet and exercise.

### And finally...

We hope this letter will help you feel more confident about giving up cigarettes. If you have a go, you have a real chance of succeeding.

With best wishes,

The Health Centre.

| day | hour | wind direction | wind speed (10m altitude) | wind speed (50m alt) |
|---|---|---|---|---|
| 12-06-02 | 6 | WSW | 10 | 12 |
| 12-06-02 | 9 | WSW | 9 | 11 |
| 12-06-02 | 12 | WSW | 7 | 9 |
| 12-06-02 | 15 | WSW | 7 | 9 |
| 12-06-02 | 18 | SW | 7 | 9 |
| 12-06-02 | 21 | SSW | 8 | 10 |
| 13-06-02 | 0 | SSW | 10 | 12 |

Figure 3: Wind data extract from 12-Jun-2002 numerical weather prediction

Knowledge acquisition in STOP was primarily based on structured expert-oriented KA techniques, including in particular sorting and think-aloud protocols. Knowledge was acquired from five health professionals; three doctors, a nurse, and a health psychologist. These experts were knowledgeable about smoking and about patient information, but they were not experts on writing tailored smoking-cessation letters. In fact there are no experts at this task, since no one manually writes tailored smoking-cessation letters.

It is not unusual for an NLG system to attempt a task which is not currently performed by human experts; other examples include descriptions of software models (Lavoie, Rambow, & Reiter, 1997), customised descriptions of museum items (Oberlander, O'Donnell, Knott, & Mellish, 1998), and written feedback for adult literacy students (Williams, Reiter, & Osman, 2003). Knowledge validation in STOP was mostly based on feedback from users (smokers), and on the results of the clinical trial.

### 2.4 SumTime-Mousam

SUMTIME-MOUSAM (Sripada, Reiter, Hunter, & Yu, 2002) is an NLG system that generates marine weather forecasts for offshore oil rigs, from numerical weather simulation data. An extract from SUMTIME-MOUSAM's input data is shown in Figure 3, and an extract from the forecast generated from this data is shown in Figure 4. From a KA perspective, the main knowledge needed by SUMTIME-MOUSAM was again what content and expression was best for users; for example,

- What changes in a meteorological parameter are significant enough to be reported in the text? The forecast in Figure 4, for example, mentions changes in wind direction but not changes in wind speed.

- What words and phrases should be used to communicate time? For example, should 1800 be described as *early evening* (as in Figure 4) or as *late afternoon*?

SUMTIME-MOUSAM is currently being used operationally by a meteorological company, to generate draft forecasts which are post-edited by human forecasters.

Knowledge acquisition in SUMTIME-MOUSAM was based on both corpus analysis of manually-written forecasts and structured KA with expert meteorologists. Unlike the experts we worked with in STOP, the meteorologists we worked with in SUMTIME-MOUSAM

---

FORECAST 6 - 24 GMT, Wed 12-Jun 2002
  WIND(KTS)

|  | 10M: | WSW 8-13 gradually backing SW by early evening and SSW by midnight. |
|  | 50M: | WSW 10-15 gradually backing SW by early evening and SSW by midnight. |

  WAVES(M)

|  | SIG HT: | 0.5-1.0 mainly SW swell. |
|  | MAX HT: | 1.0-1.5 mainly SW swell. |
|  | PER(SEC) |  |

               WAVE PERIOD: Wind wave 3-5 mainly 6 second SW swell.
               WINDWAVE PERIOD: 3-5.
               SWELL PERIOD: 5-7.

WEATHER:        Partly cloudy becoming overcast with light rain around midnight.
VIS(NM):          Greater than 10 reduced to 5-8 in precipitation.
AIR TEMP(C):    8-10 rising 9-11 around midnight.
CLOUD(OKTAS/FT): 2-4 CU/SC 1300-1800 lowering 7-8 ST/SC 700-900 around midnight.

---

Figure 4: Extract from forecast generated for 12-Jun-2002

were experienced at writing the target texts (weather forecasts). The forecast corpus included the numerical weather simulation data that the forecasters used when writing the forecasts, as well as the actual forecast texts (Sripada, Reiter, Hunter, & Yu, 2003).

Knowledge validation in SUMTIME-MOUSAM has mostly been conducted by checking knowledge acquired from the corpus with the experts, and checking knowledge acquired from the experts against the corpus. In other words, we have tried to make the validation technique as different as possible from the acquisition technique. We are currently evaluating SUMTIME-MOUSAM as a system by measuring the number of edits that forecasters make to the computer-generated draft forecasts.

## 3. Knowledge Acquisition Techniques Tried

In this section we summarise the main KA techniques we used in STOP and SUMTIME-MOUSAM. For each technique we give an example of the knowledge acquired, and discuss what we learned when we tried to validate the knowledge. Table 1 gives a very high level overview of the major advantages and disadvantages of the different techniques we tried, when the different techniques were perhaps most useful, and what types of knowledge they were best suited to acquiring (using the classification of Section 2.1). As this table shows, no one technique is clearly best; they all have different strengths and weaknesses. Probably the best overall KA strategy is to use a mix of different techniques; we will further discuss this in Section 5.

| Techniques | Advantages | Disadvantages | When Useful | Types of Knowledge |
|---|---|---|---|---|
| directly ask experts | get big picture | many gaps, may not match practice | initial prototype | domain, DCK |
| structured KA with experts | get details, get rationale | limited coverage, experts variable | flesh out prototype | depends on expert |
| corpus analysis | get lots of knowledge quickly | hard to create, texts inconsistent, poor models for NLG | robustness, unusual cases | DCK, communication |
| expert revision | fix problems in knowledge | local optimisation, not major changes | improve system | all |

Table 1: Summary Evaluation of KA techniques for NLG

## 3.1 Directly Asking Experts for Knowledge

The simplest and perhaps most obvious KA technique for NLG is to simply ask experts how to write the texts in question. In both STOP and SumTime-Mousam, experts initially gave us spreadsheets or flowcharts describing how they thought texts should be generated. In both projects, it also turned out that the experts' description of how texts should be generated did not in fact match how people actually wrote the texts in question. This is a common finding in KA, and it is partially due to the fact that it is difficult for experts to introspectively examine the knowledge they use in practice (Anderson, 1995); this is why proponents of expert-oriented KA prefer structured KA techniques.

For example, at the beginning of SumTime-Mousam, one of the meteorologists gave us a spreadsheet which he had designed, which essentially encoded how he thought some parts of weather forecasts should be generated (the spreadsheet did not generate a complete weather forecast). We analysed the logic used in the spreadsheet, and largely based the first version of SumTime-Mousam on this logic.

One goal of our analysis was to create an algorithm that could decide when a change in a parameter value was significant enough so that it should be mentioned in the weather report. The spreadsheet used context-dependent change thresholds to make this decision. For example, a change in the wind speed would be mentioned if

- the change was 10 knots or more, and the final wind speed was 15 knots or less;

- the change was 5 knots or more, and the final wind speed was between 15 and 40 knots; or

- the change was 10 knots or more, and the final wind speed was over 40 knots.

The context-dependent thresholds reflect the usage of the weather reports by the users (in this case, oil company staff making decisions related to North Sea offshore oil rigs). For example, if a user is deciding how to unload a supply boat, moderate changes in wind speed don't matter at low speeds (because light winds have minimal impact on supply boat operations) and at high speeds (because the boat won't even attempt to unload in very heavy winds), but may affect decisions at in-between speeds. The context-dependent thresholds would be expected to vary according to the specific forecast recipient, and should be set in consultation with the recipient.

From our perspective, there were two main pieces of knowledge encoded in this algorithm:

1. The absolute size of a change determines whether it should be mentioned or not, and

2. The threshold for significance depends on the context and ultimately on how the user will use the information.

### 3.1.1 Validation of Direct Expert Knowledge

We checked these rules by comparing them to what we observed in our corpus analysis of manually written forecasts (Section 3.3). This suggested that while (2) above is probably correct, (1) may be incorrect. In particular, a linear segmentation model (Sripada et al., 2002), which basically looks at changes in slope rather than changes in the absolute value of a parameter, better matches the corpus texts. The expert who designed the spreadsheet model agreed that segmentation was probably a better approach. He also essentially commented that one reason for his use of the absolute size model was that this was something that was easily comprehensible to someone who was neither a programmer nor an expert at numerical data analysis techniques.

In other words, in addition to problems in introspecting knowledge, it also perhaps is not reasonable to expect a domain expert to be able to write a sophisticated data analysis algorithm based on his expertise. This is not an issue if the knowledge needed is purely declarative, as it is in many AI applications; but if we need procedural or algorithmic knowledge, we must bear in mind that domain experts may not have sufficient computational expertise to express their knowledge as a computer algorithm.

### 3.1.2 Role of Directly Asking Experts for Knowledge

Although the expert's spreadsheet in SumTime-Mousam was far from ideal, it was extremely useful as a starting point. It specified an initial system which we could build fairly easily, and which produced at least vaguely plausible output. Much the same in fact happened in STOP, when one of the doctors gave us a flowchart which certainly had many weaknesses, but which was useful as an initial specification of a relatively easy-to-build and somewhat plausible system. In both STOP and SumTime-Mousam, as indeed in other NLG projects we have been involved in, having an initial prototype system working as soon as possible was very useful for developing our ideas and for explaining to domain experts and other interested parties what we were trying to do.

In terms of the types of knowledge mentioned in Section 2.1, both the STOP flowchart and the SumTime-Mousam spreadsheet specified domain knowledge (for example, how smokers should be categorised) and domain communication knowledge (for example, the use of ranges instead of single numbers to communicate wind speed). The STOP flowchart did not specify any generic communication knowledge such as English grammar and morphology; the author probably believed we knew more about such things than he did. The SumTime-Mousam spreadsheet did in effect include a few English grammar rules, but these were just to get the spreadsheet to work, the author did not have much confidence in them.

In summary, we think directly asking experts for knowledge is an excellent way to quickly build an initial system, especially if the NLG developers can supply communication knowledge that the domain expert may not possess. But once the initial system is in place, it is probably best to use other KA techniques, at least in poorly understood areas such as NLG. However, in applications where there is a solid theoretical basis, and the expert can simply say 'build your system according to theory X', an expert's direct knowledge may perhaps be all that is needed.

## 3.2 Structured Expert-Oriented KA: Think-Aloud Protocols

There are numerous types of structured expert-oriented KA techniques, including think-aloud protocols, sorting, and structured interviews (Scott et al., 1991). We will focus here on think-aloud protocols, which is the technique we have used the most. We have tried other structured KA techniques as well, such as sorting (Reiter et al., 2000); we will not describe these here, but our broad conclusions about other structured KA techniques were similar to our conclusions about think-aloud protocols.

In a think-aloud protocol, an expert carries out the task in question (in our case, writing a text) while 'thinking aloud' into an audio (or video) recorder. We used think-aloud protocols in both STOP and SumTime-Mousam. They were especially important in STOP, where they provided the basis for most content and phrasing rules.

A simple example of the think-aloud process is as follows. One of the doctors wrote a letter for a smoker who had tried to stop before, and managed to stop for several weeks before starting again. The doctor made the following comments in the think-aloud transcript:

> Has he tried to stop smoking before? Yes, and the longest he has managed to stop — he has ticked the one week right up to three months and that's encouraging in that he has managed to stop at least once before, because it is always said that the people who have had one or two goes are more likely to succeed in the future.

He also included the following paragraph in the letter that he wrote for this smoker:

> I see that you managed to stop smoking on one or two occasions before but have gone back to smoking, but you will be glad to know that this is very common and most people who finally stop smoking have had one or two attempts in the past before they finally succeed. What it does show is that you are capable of stopping even for a short period, and that means you are much more likely to be able to stop permanently than somebody who has never ever stopped smoking at all.

After analysing this session, we proposed two rules:

- IF (previous attempt to stop) THEN (message: more likely to succeed)

- IF (previous attempt to stop) THEN (message: most people who stop have a few unsuccessful attempts first)

The final system incorporated a rule (based on several KA sessions, not just the above one) that stated that if the smoker had tried to stop before, and if the letter included a section on confidence building, then the confidence-building section should include a short message about previous attempts to stop. If the smoker had managed to quit for more than one week, this should be mentioned in the message; otherwise the message should mention the recency of the smoker's previous cessation attempt if this was within the past 6 months. The actual text generated from this rule in the example letter of Figure 2 is

> Although you don't feel confident that you would be able to stop if you were to try, you have several things in your favour.
>
> • You have stopped before for more than a month.

Note that the text produced by the actual STOP code is considerably simpler than the text originally written by the expert. This is fairly common, as are simplifications in the logic used to decide whether to include a message in a letter or not. In many cases this is due to the expert having much more knowledge and expertise than the computer system (Reiter & Dale, 2000, pp 30–36). In general, the process of deriving implementable rules for NLG systems from think-aloud protocols is perhaps more of an art than a science, not least because different experts often write texts in very different ways.

### 3.2.1 Validation of Structured KA Knowledge

We attempted to verify some of the rules acquired from STOP think-aloud sessions by performing a series of small experiments where we asked smokers to comment on a letter, or to compare two versions of a letter. Many of the rules were supported by these experiments - for example, people in general liked the recap of smoking likes and dislikes (see **You have good reasons to stop...** section of Figure 2). However, one general negative finding of these experiments was that the tailoring rules were insufficiently sensitive to unusual or atypical aspects of individual smokers; and most smokers were probably unusual or atypical in some way. For example, STOP letters did not go into the medical details of smoking (as none of the think-aloud expert-written letters contained such information), and while this seemed like the right choice for many smokers, a few smokers did say that they would have liked to see more medical information about smoking. Another example is that (again based on the think-aloud sessions) we adopted a positive tone and did not try to scare smokers; and again this seemed right for most smokers, but some smokers said that a more 'brutal' approach would be more effective for them.

The fact that our experts did not tailor letters in such ways may possibly reflect the fact that such tailoring would not have been appropriate for the relatively small number of specific cases they considered in our think-aloud sessions. We had 30 think-aloud sessions with experts, who looked at 24 different smoker questionnaires (6 questionnaires were considered by two experts). This may sound like a lot, but it is a drop in the ocean when we consider how tremendously variable people are.

Comments made by smokers during the STOP clinical trial (Reiter, Robertson, & Osman, 2003) also revealed some problems with think-aloud derived rules. For example, we decided not to include practical 'how-to-stop' information in letters for people not currently intending to stop smoking; smoker comments suggest that this was a mistake. In fact, some experts did include such information in think-aloud letters for such people, and some did not. Our decision not to include this information was influenced by the Stages of Change theoretical model (Prochaska & diClemente, 1992) of behaviour change, which states that 'how-to-stop' advice is inappropriate for people not currently intending to stop; in retrospect, this decision was probably a mistake.

We repeated two of our think-aloud exercises 15 months after we originally performed them; that is, we went back to one of our experts and gave him two questionnaires he had analysed 15 months earlier, and asked him to once again think aloud while writing letters based on the questionnaires. The letters that the expert wrote in the second session were somewhat different from the ones he had originally written, and were preferred by smokers over the letters he had originally written (Reiter et al., 2000). This suggests that our experts were not static knowledge sources, but were themselves learning about the task of writing tailored smoking-cessation letters during the course of the project. Perhaps this should not be a surprise given that none of the experts had ever attempted to write such letters before getting involved with our project.

### 3.2.2 Role of Structured Expert-Oriented KA

Structured expert-oriented KA was certainly a useful way to expand, refine, and generally improve initial prototypes constructed on the basis of experts' direct knowledge. By focusing on actual cases and by structuring the KA process, we learned many things which the experts did not mention directly. We obtained all the types of knowledge mentioned in Section 2.1, by working with experts with the relevant expertise. For example in STOP we acquired domain knowledge (such as the medical effects of smoking) from doctors, domain communication knowledge (such as which words to use) from a psychologist with expertise in writing patient information leaflets, and communication knowledge about graphic design and layout from a graphic designer.

However, structured expert-oriented KA did have some problems, including in particular coverage and variability. As mentioned above, 30 sessions that examined 24 smoker questionnaires could not possibly give good coverage of the population of smokers, given how complex and variable people are. As for variation, the fact that different experts wrote texts in very different ways made it difficult to extract rules from the think-aloud protocols. We undoubtedly made some mistakes in this regard, such as not giving 'how-to-stop' information to people not currently intending to stop smoking. Perhaps we should have focused on a single expert in order to reduce variation. However, our experiences suggested that different experts were better at different types of information, and also that experts changed over time (so we might see substantial variation even in texts from a single author); these observations raise doubts about the wisdom and usefulness of a single-expert strategy.

In short, the complexity of NLG tasks means that a very large number of structured KA sessions may be needed to get good coverage; and the fact that there are numerous ways to write texts to fulfill a communicative goal means that different experts tend to write very differently, which makes analysis of structured KA sessions difficult.

## 3.3 Corpus Analysis

In recent years there has been great interest in Natural Language Processing and other areas of AI in using machine learning techniques to acquire knowledge from relevant data sets. For example, instead of building a medical diagnosis system by trying to understand how expert doctors diagnose diseases, we can instead analyse data sets of observed symptoms and actual diseases, and use statistical and machine learning techniques to determine which symptoms predict which disease. Similarly, instead of building an English grammar by working with expert linguists, we can instead analyse large collections of grammatical English texts in order to learn the allowable structures (grammar) of such texts. Such collections of texts are called *corpora* in Natural Language Processing.

There has been growing interest in applying such techniques to learn the knowledge needed for NLG. For example, Barzilay and McKeown (2001) used corpus-based machine learning to learn paraphrase possibilities; Duboue and McKeown (2001) used corpus-based machine learning to learn how NP constituents should be ordered; and Hardt and Rambow (2001) used corpus-based machine learning to learn rules for VP ellipsis.

Some NLG researchers, such as McKeown et al. (1994), have used the term 'corpus analysis' to refer to the manual analysis (without using machine learning techniques) of a small set of texts which are written explicitly for the NLG project by domain experts (and hence are not naturally occurring). This is certainly a valid and valuable KA technique, but we regard it as a form of structured expert-oriented KA, in some ways similar to think-aloud protocols. In this paper, 'corpus analysis' refers to the use of machine learning and statistical techniques to analyse collections of naturally occurring texts.

Corpus analysis in our sense of the word was not possible in STOP because we did not have a collection of naturally occurring texts (since doctors do not currently write personalised smoking-cessation letters). We briefly considered analysing the example letters produced in the think-aloud sessions with machine learning techniques, but we only had 30 such texts, and we believed this would be too few for successful learning, especially given the high variability between experts. In other words, perhaps the primary strength of corpus analysis is its ability to extract information from large data sets; but if there are no large data sets to extract information from, then corpus analysis loses much of its value.

In SumTime-Mousam, we were able to acquire and analyse a substantial corpus of 1099 human-written weather forecasts, along with the data files that the forecasters looked at when writing the forecasts (Sripada et al., 2003). Details of our corpus analysis procedures and results have been presented elsewhere (Reiter & Sripada, 2002a; Sripada et al., 2003), and will not be repeated here.

### 3.3.1 Validation of Corpus Analysis Knowledge

While many of the rules we acquired from corpus analysis were valid, some rules were problematical, primarily due to two factors: individual variations between the writers, and writers making choices that were appropriate for humans but not for NLG systems.

A simple example of individual variation and the problems it causes is as follows. One of the first things we attempted to learn from the corpus was how to express numbers in wind statements. We initially did this by searching for the most common textual realisation of each number. This resulted in rules that said that 5 should be expressed as *5*, but 6 should be expressed as *06*. Now it is probably acceptable for a forecast to always include leading zeros for single digits (that is, use *05* and *06*), and to never include leading zeros (that is, use *5* and *6*). However, it is probably not acceptable to mix the two (that is, use *5* and *06* in the same forecast), which is what our rules would have led to.

| form | F1 | F2 | F3 | F4 | F5 | unknown | total |
|------|----|----|-----|-----|-----|---------|-------|
| *5*  | 0  | 7  | 0   | 0   | 122 | 4       | 133   |
| *05* | 0  | 0  | 1   | 46  | 0   | 2       | 49    |
| *6*  | 0  | 44 | 0   | 0   | 89  | 2       | 135   |
| *06* | 0  | 0  | 364 | 154 | 0   | 13      | 531   |

Table 2: Usage of *5*, *05*, *6*, *06* in wind statements, by forecaster

The usage of *5*, *05*, *6*, and *06* by each individual forecaster is shown in Table 2. As this table suggests, each individual forecaster is consistent; forecasters F3 and F4 always include leading zeros, while forecasters F2 and F5 never include leading zeros. F1 in fact is also consistent and always omits leading zeros; for example he uses *8* instead of *08*. The reason that the overall statistics favour *5* over *05* but *06* over *6* is that individuals also differ in which descriptions of wind speed they prefer to use. For example, F1 never explicitly mentions low wind speeds such as 5 or 6 knots, and instead always uses generic phrases such as *10 OR LESS*; F2, F4, and F5 use a mix of generic phrases and explicit numbers for low wind speeds; and F3 always uses explicit numbers and never uses generic phrases. Some of the forecasters (especially F3) also have a strong preference for even numbers. This means that the statistics for *5* vs. *05* are dominated by F5 (the only forecaster who both explicitly mentions low wind speeds and does not prefer even numbers); while the statistics for *6* vs. *06* are dominated by F3 (who uses this number a lot because he avoids both generic phrases and odd numbers). Hence the somewhat odd result that the corpus overall favours *5* over *05* but *06* over *6*.

This example is by no means unique. Reiter and Sripada (2002b) explain how a more complex analysis using this corpus, whose goal was to determine the most common time phrase for each time, similarly led to unacceptable rules, again largely because of individual differences between the forecasters.

There are obvious methods to deal with the problems caused by individual variation. For example, we could restrict the corpus to texts from one author; although this does have the major drawback of significantly reducing the size of the corpus. We could also use a more sophisticated model, such as learning one rule for how all single digit numbers are expressed, not separate rules for each number. Or we could analyse the behaviour of individuals and identify choices (such as presence of a leading zero) that vary between individuals but are consistently made by any given individual; and then make such choices parameters which the user of the NLG system can specify. This last option is probably the best for NLG systems (Reiter, Sripada, & Williams, 2003), and is the one used in SumTime-Mousam for the leading-zero choice.

Our main point is simply that we would have been in trouble if we had just accepted our initial corpus-derived rules (use *5* and *06*) without question. As most corpus researchers are of course aware, the result of corpus analysis depends on what is being learned (for example, a rule on how to realise 5, or a rule on how to realise all single-digit numbers) and on what features are used in the learning (for example, just the number, or the number and the author). In more complex analyses, such as our analysis of time-phrase choice rules (Reiter & Sripada, 2002b), the result also depends on the algorithms used for learning and alignment. The dependence of corpus analysis on these choices means that the results of a particular analysis are not guaranteed to be correct and need to be validated (checked) just like the results of other KA techniques. Also, what is often the best approach from an NLG perspective, namely identifying individual variations and letting the user choose which variation he or she prefers, requires analysing differences between individual writers. To the best of our knowledge most published NL corpus analyses have not done this, perhaps in part because many popular corpora do not include author information.

The other recurring problem with corpus-derived rules was cases where the writers produced sub-optimal texts that in particular were shorter than they should have been, probably because such texts were quicker to write. For instance, we noticed that when a parameter changed in a more or less steady fashion throughout a forecast period, the forecasters often omitted a time phrase. For example, if a S wind rose steadily in speed from 10 to 20 over the course of a forecast period covering a calendar day, the forecasters might write S 8-12 RISING TO 18-22, instead of S 8-12 RISING TO 18-22 BY MIDNIGHT. A statistical corpus analysis showed that the 'null' time phrase was the most common one in such contexts, used in 33% of cases. The next most common time phrase, *later*, was only used in 14% of cases. Accordingly, we programmed our system to omit the time phrase in such circumstances. However, when we asked experts to comment on and revise our generated forecasts (Section 3.4), they told us that this behaviour was incorrect, and that forecasts were more useful to end users if they included explicit time phrases and did not rely on the readers remembering when forecast periods ended. In other words, in this example the forecasters were doing the wrong thing, which of course meant that the rule produced by corpus analysis was incorrect.

We don't know why the forecasters did this, but discussions with the forecast managers about this and other mistakes (such as forecast authors describing wind speed and direction as changing at the same time, even when they actually were predicted to change at different times) suggested that one possible cause is the desire to write forecasts quickly. In particular, numerical weather predictions are constantly being updated, and customers want their forecasts to be based on the most up-to-date prediction; this can limit the amount of time available to write forecasts.

In fact it can be perfectly rational for human writers to 'cut corners' because of time limitations. If the forecasters believe, for example, that quickly writing a forecast at the last minute will let them use more up-to-date prediction data; and that the benefits of more up-to-date data outweighs the costs of abbreviated texts, then they are making the right decision when they write shorter-than-optimal texts. An NLG system, however, faces a very different set of tradeoffs (for example, omitting a time phrase is unlikely to speed up an NLG system), which means that it should not blindly imitate the choices made by human writers.

This problem is perhaps a more fundamental one than the individual variation problem, because it can not be solved by appropriate choices as to what is being learned, what features are considered, and so forth. Corpus analysis, however it is performed, learns the choice rules used by human authors. If these rules are inappropriate for an NLG system, then the rules learned by corpus analysis will be inappropriate ones as well, regardless of how the corpus analysis is carried out.

In very general terms, corpus analysis certainly has many strengths, such as looking at what people do in practice, and collecting large data sets which can be statistically analysed. But pure corpus analysis does perhaps suffer from the drawback that it gives no information on why experts made the choices they made, which means that blindly imitating a corpus can lead to inappropriate behaviour when the human writers face a different set of constraints and tradeoffs than the NLG system.

### 3.3.2 Role of Corpus Analysis

Corpus analysis and machine learning are wonderful ways to acquire knowledge if

1. there is a large data set (corpus) that covers unusual and boundary cases as well as normal cases;

2. the members of the data set (corpus) are correct in that they are what we would like the software system to produce; and

3. the members of the data set (corpus) are consistent (modulo some noise), for example any given input generally leads to the same output.

These conditions are probably satisfied when learning rules for medical diagnosis or speech recognition. However, they were not satisfied in our projects. None of the above conditions were satisfied in STOP, and only the first was satisfied in SumTime-Mousam.

Of course, there may be ways to alleviate some of these problems. For example, we could try to acquire general communication knowledge which is not domain dependent (such as English grammar) from general corpora such as the British National Corpus; we could argue that certain aspects of manually written texts (such as lexical usage) are unlikely to be adversely affected by time pressure and hence are probably correct; and we could analyse the behaviour of individual authors in order to enhance consistency (in other words, treat author as an input feature on a par with the actual numerical or semantic input data). There is scope for valuable research here, which we hope will be considered by people interested in corpus-based techniques in NLG.

We primarily used corpus analysis in SumTime-Mousam to acquire domain communication knowledge, such as how to linguistically express numbers and times in weather forecasts, when to elide information, and sublanguage constraints on the grammar of our weather forecasts. Corpus analysis of course can also be used to acquire generic communication knowledge such as English grammar, but as mentioned above this is probably best done on a large general corpus such as the British National Corpus. We did not use corpus analysis to acquire domain knowledge about meteorology. Meteorological researchers in fact do use machine learning techniques to learn about meteorology, but they analyse numeric data sets of actual and predicted weather, they do not analyse textual corpora.

In summary, machine learning and corpus-based techniques are extremely valuable if the above conditions are satisfied, and in particular offer a cost-effective solution to the problem of acquiring the large amount of knowledge needed in complex NLG applications (Section 3.2.2). Acquiring large amounts of knowledge using expert-oriented KA techniques is expensive and time-consuming because it requires many sessions with experts; in contrast, if a large corpus of consistent and correct texts can be created, then large amounts of knowledge can be extracted from it at low marginal cost. But like all learning techniques, corpus analysis is very vulnerable to the 'Garbage In, Garbage Out' principle; if the corpus is small, incorrect, and/or inconsistent, then the results of corpus analysis may not be correct.

## 3.4 Expert Revision

In both STOP and SUMTIME-MOUSAM, we made heavy use of expert revision. That is, we showed generated texts to experts and asked them to suggest changes that would improve them. In a sense, expert revision could be considered to be a type of structured expert-oriented KA, but it seems to have somewhat different strengths and weaknesses than the techniques mentioned in Section 3.2, so we treat it separately.

As an example of expert revision, an early version of the STOP system used the phrase *there are lots of good reasons for stopping*. One of the experts commented during a revision session that the phrasing should be changed to emphasise that the reasons listed (in this particular section of the STOP letter) were ones the smoker himself had selected in the questionnaire he filled out. This eventually led to the revised wording *It is encouraging that you have many good reasons for stopping*, which is in the first paragraph of the example letter in Figure 2. An example of expert revision in SUMTIME-MOUSAM was mentioned in Section 3.3; when we showed experts generated texts that omitted some end-of-period time phrases, they told us this was incorrect, and we should include such time phrases.

In STOP, we also tried revision sessions with recipients (smokers). This was less successful than we had hoped. Part of the problem was the smokers knew very little about STOP (unlike our experts, who were all familiar with the project), and often made comments which were not useful for improving the system, such as *I did stop for 10 days til my daughter threw a wobbly and then I wanted a cigarette and bought some*. Also, most of our comments came from well-educated and articulate smokers, such as university students. It was harder to get feedback from less well-educated smokers, such as single mothers living in council (public housing) estates. Hence we were unsure if the revision comments we obtained were generally applicable or not.

### 3.4.1 VALIDATION OF EXPERT REVISION KNOWLEDGE

We did not validate expert revision knowledge as we did with the other techniques. Indeed, we initially regarded expert revision as a validation technique, not a KA technique, although in retrospect it probably makes more sense to think of it as a KA technique.

On a qualitative level, though, expert revision has certainly resulted in a lot of useful knowledge and ideas for changing texts, and in particular proved a very useful way of improving the handling of unusual and boundary cases. For example, we changed the way we described uneventful days in SUMTIME-MOUSAM (when the weather changed very little during a day) based on revision sessions.

The comment was made during STOP that revision was best at suggesting specific localised changes to generated text, and less useful in suggesting larger changes to the system. One of the STOP experts suggested, after the system was built, that he might have been able to suggest larger changes if we had explained the system's reasoning to him, instead of just giving him a letter to revise. In other words, just as we asked experts to 'think-aloud' as they wrote letters, in order to understand their reasoning, it could be useful in revision sessions if experts understood what the computer system was 'thinking' as well as what it actually produced. Davis and Lenat (1982, page 260) have similarly pointed out that explanations can help experts debug and improve knowledge-based systems.

### 3.4.2 Role of Expert Revision

We have certainly found expert revision to be an extremely useful technique for improving NLG systems; and furthermore it is useful for improving all types of knowledge (domain, domain communication, and communication). But at the same time revision does seem to largely be a local optimisation technique. If an NLG system is already generating reasonable texts, then revision is a good way of adjusting the system's knowledge and rules to improve the quality of generated text. But like all local optimisation techniques, expert revision may tend to push systems towards a 'local optimum', and may be less well suited to finding radically different solutions that give a better result.

## 4. Discussion: Problems Revisited

In section 1 we explained that writing tasks can be difficult to automate because these are complex, often novel, poorly understood, and allow multiple solutions. In this section we discuss each of these problems in more detail, based on our experiences with STOP and SumTime-Mousam.

### 4.1 Complexity

Because NLG systems communicate with humans, they need knowledge about people, language, and how people communicate; since all of these are very complex, that means that in general NLG systems need a lot of complex knowledge. This is one of the reasons why knowledge acquisition for NLG is so difficult. If we recall the distinction in Section 2.1 between domain knowledge, domain communication knowledge, and communication knowledge, it may be that communication knowledge (such as grammar) is generic and hence can be acquired once (perhaps by corpus-based techniques) and then used in many applications. And domain knowledge is similar to what is needed by other AI systems, so problems acquiring it are not unique to NLG. But domain communication knowledge, such as the optimal tone of a smoking letter and how this tone can be achieved, or when information in a weather forecast can be elided, is application dependent (and hence cannot be acquired generically) and is also knowledge about language and communication (and hence is complex). Hence KA for NLG may always require acquiring complex knowledge.

In our experience, the best way to acquire complex knowledge robustly is to get information on how a large number of individual cases are handled. This can be done by corpus analysis if a suitable corpus can be created. It can also sometimes be done by expert revision, if experts have the time to look at a large number of generated texts; in this regard it may be useful to tell them to only comment on major problems and to ignore minor difficulties. But however the knowledge is acquired, it will require a substantial effort.

## 4.2 Novelty

Of course, many AI systems need complex knowledge, so the above comments are hardly unique to NLG. But one aspect of NLG which perhaps is more unusual is that many of the tasks NLG systems are expected to perform are novel tasks that are not currently done by humans. Most AI 'expert systems' attempt to replicate the performance of human experts in areas such as medical diagnosis and credit approval. Similarly, most language technology systems attempt to replicate the performance of human language users in tasks such as speech recognition and information retrieval. But many NLG applications are like STOP, and attempt a task that no human performs. Even in SUMTIME-MOUSAM, an argument could be made that the task humans actually perform is writing weather forecasts under time constraints, which is in fact different from the task performed by SUMTIME-MOUSAM.

Novelty is a fundamental problem, because it means that knowledge acquired from expert-oriented KA may not be reliable (since the experts are not in fact experts at the actual NLG task), and that a corpus of manually-written texts probably does not exist. This means that none of the KA techniques described above are likely to work. Indeed, acquiring novel knowledge is almost the definition of scientific research, so perhaps the only way to acquire such knowledge is to conduct scientific research in the domain. Of course, only some knowledge will need to be acquired in this way, even in a novel application it is likely that much of the knowledge needed (such as grammar and morphology) is not novel.

On the other hand, novelty perhaps is also an opportunity for NLG. One of the drawbacks of conventional expert systems is that their performance is often limited to that of human experts, in which case users may prefer to consult actual experts instead of computer systems. But if there are no experts at a task, an NLG system may be used even if its output is far from ideal.

## 4.3 Poorly Understood Tasks

A perhaps related problem is that there are no good theoretical models for many of the choices that NLG systems need to make. For example, the ultimate goal of STOP is to change people's behaviour, and a number of colleagues have suggested that we base STOP on argumentation theory, as Grasso, Cawsey, and Jones (2000) did for their dietary advice system. However, argumentation theory focuses on persuading people to change their beliefs and desires, whereas the goal of STOP was more to encourage people to act on beliefs and desires they already had. In other words, STOP's main goal was to encourage people who already wanted to stop smoking to make a serious cessation attempt, not to convince people who had no desire to quit that they should change their mind about the desirability of smoking. The most applicable theory we could find was Stages of Change (Prochaska & diClemente, 1992), and indeed we partially based STOP on this theory. However, the results of our evaluation suggested that some of the choices and rules that we based on Stages of Change were incorrect, as mentioned in Section 3.2.1.

Similarly, one of the problems in SUMTIME-MOUSAM is generating texts that will be interpreted correctly despite the fact that different readers have different idiolects and in particular probably interpret words in different ways (Reiter & Sripada, 2002a; Roy, 2002). Theoretical guidance on how to do this would have been very useful, but we were not able to find any such guidance.

The lack of good theoretical models means that NLG developers cannot use such models to 'fill in the cracks' between knowledge acquired from experts or from data sets, as can be done by AI systems in better understood areas such as scheduling or configuring machinery. This in turn means that a lot of knowledge must be acquired. In applications where there is a good theoretical basis, the goal of KA is perhaps to acquire a limited amount of high-level information about search strategies, taxonomies, the best way to represent knowledge, etc; once these have been determined, the details can be filled in by theoretical models. But in applications where details cannot be filled in from theory and need to be acquired, much more knowledge is needed. Acquiring such knowledge with structured expert-oriented KA could be extremely expensive and time consuming. Corpus-based techniques are cheaper if a large corpus is available; however, the lack of a good theoretical understanding perhaps contributes to the problem that we do not know which behaviour we observe in the corpus is intended to help the reader (and hence should be copied by an NLG system) and which behaviour is intended to help the writer (and hence perhaps should not be copied).

### 4.4 Expert Variation

Perhaps in part because of the lack of good theories, in both STOP and SumTime-Mousam we observed considerable variation between experts. In other words, different experts wrote quite different texts from the same input data. In STOP we also discovered that experts changed how they wrote over time (Section 3.2.1).

Variability caused problems for both structured expert-oriented KA (because different experts told us different things) and for corpus analysis (because variation among corpus authors made it harder to extract a consistent set of rules with good coverage). However, variation seems to have been less of a problem with revision. We suspect this is because experts vary less when they are very confident about a particular decision; and in revision experts tended to focus on things they were confident about, which was not the case with the other KA techniques.

In a sense variability may be especially dangerous in corpus analysis, because there is no information in a corpus about the degree of confidence authors have in individual decisions, and also because developers may not even realise that there is variability between authors, especially if the corpus does not include author information. In contrast, structured expert-oriented techniques such as think-aloud do sometimes give information about experts' confidence, and also variations between experts are usually obvious.

We experimented with various techniques for resolving differences between experts/authors, such as group discussions and focusing on the decisions made by one particular expert. None of these were really satisfactory. Given our experiences with revision, perhaps the best way to reduce variation is to develop KA techniques that very clearly distinguish between decisions experts are confident in and decisions they have less confidence in.

## 5. Development Methodology: Using Multiple KA Techniques

From a methodological perspective, the fact that different KA techniques have different strengths and weaknesses suggests that it makes sense to use a mixture of several different KA techniques. For example, if both structured expert-oriented KA and corpus analysis are used, then the explanatory information from the expert-oriented KA can be used to help identify which decisions are intended to help the reader and which are intended to help the writer, thus helping overcome a problem with corpus analysis; and the broader coverage of corpus analysis can show how unusual and boundary cases should be handled, thus overcoming a problem with expert-oriented KA.

It also may make sense to use different techniques at different points in the development process. For example, directly asking experts for knowledge could be stressed at the initial stages of a project, and used to build a very simple initial prototype; structured KA with experts and corpus analysis could be stressed during the middle phases of a project, when the prototype is fleshed out and converted into something resembling a real system; and revision could be used in the later stages of a project, when the system is being refined and improved.

This strategy, which is graphically shown in Figure 5, is basically the one we followed in both STOP and SUMTIME-MOUSAM. Note that it suggests that knowledge acquisition is something that happens throughout the development process. In other words, we do not first acquire knowledge and then build a system; knowledge acquisition is an ongoing process which is closely coupled with the general software development effort. Of course, this is hardly a novel observation, and there are many development methodologies for knowledge-based systems that stress iterative development and continual KA (Adelman & Riedel, 1997).

In the short term, we believe that using a development methodology that combines different KA techniques in this manner, and also validating knowledge as much as possible, are the best strategies for acquiring NLG knowledge. We also believe that whenever possible knowledge that is acquired in one way should be validated in another way. In other words, we do not recommend validating corpus-acquired knowledge using corpus techniques (even if the validation is done with a held-out test set); or validating expert-acquired knowledge using expert-based validation (even if the validation is done using a different expert). It is preferable (although not always possible) to validate corpus-acquired knowledge with experts, and to validate expert-acquired knowledge with a corpus.

Another issue related to development methodology is the relationship between knowledge acquisition and system evaluation. Although these are usually considered to be separate activities, in fact they can be closely related. For example, we are currently running an evaluation of SUMTIME-MOUSAM which is based on the number of edits that forecasters manually make to computer-generated forecasts before publishing them; this is similar to edit-cost evaluations of machine translation systems (Jurafsky & Martin, 2000, page 823). However, these edits are also an excellent source of data for improving the system via expert revision. To take one recent example, a forecaster edited the computer-generated text *SSE 23-28 GRADUALLY BACKING SE 20-25* by dropping the last speed range, giving *SSE 23-28 GRADUALLY BACKING SE*. This can be considered as evaluation data (2 token edits needed to make text acceptable), or as KA data (we need to adjust our rules for eliding similar but not identical speed ranges).

In other words, real-world feedback on the effectiveness and quality of generated texts can often be used to either improve or evaluate an NLG system. How such data should be used depends on the goals of the project. In scientific projects whose goal is to test hypotheses, it may be appropriate at some point to stop improving a system and use all new effectiveness data purely for evaluation and hypothesis testing; in a sense this is analogous to holding back part of a corpus for testing purposes. In applied projects whose goal is to build a maximally useful system, however, it may be more appropriate to use all of the effectiveness data to improve the quality of the generated texts.

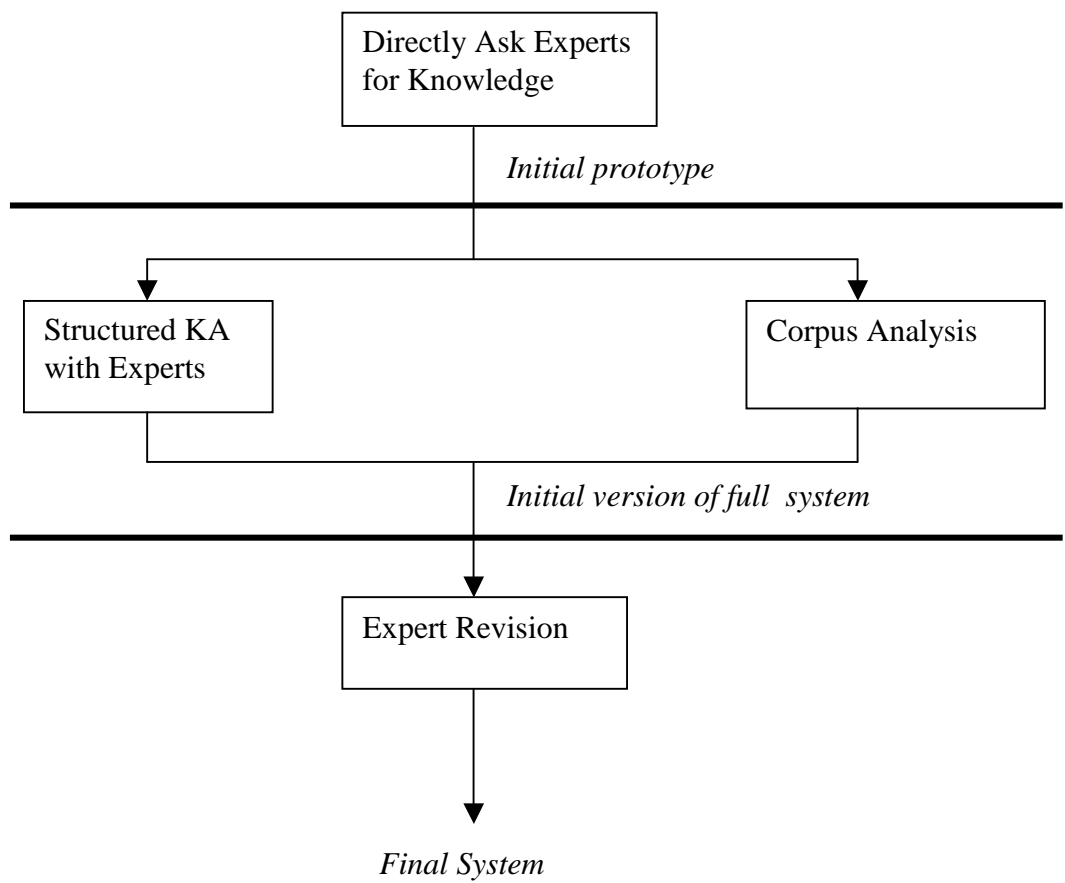

Figure 5: Our Methodology

## 6. Conclusion

Acquiring correct knowledge for NLG is very difficult, because the knowledge needed is largely knowledge about people, language, and communication, and such knowledge is complex and poorly understood. Furthermore, perhaps because writing is more of an art than a science, different people write very differently, which further complicates the knowledge acquisition process; and many NLG systems attempt novel tasks not currently done manually, which makes it very hard to find knowledgeable experts or good quality corpora. Perhaps because of these problems, every single KA technique we tried in STOP and SumTime-Mousam had major problems and limitations.

There is no easy solution to these problems. In the short term, we believe it is useful to use a mixture of different KA techniques (since techniques have different strengths and weaknesses), and to validate knowledge whenever possible, preferably using a different tech-nique than the one used to acquire the knowledge. It also helps if developers understand the weaknesses of different techniques, such as the fact that structured expert-oriented KA may not give good coverage of the complexities of people and language, and the fact that corpus-based KA does not distinguish between behaviour intended to help the reader and behaviour intended to help the writer.

In the longer term, we need more research on better KA techniques for NLG. If we cannot reliably acquire the knowledge needed by AI approaches to text generation, then there is no point in using such approaches, regardless of how clever our algorithms or models are. The first step towards developing better KA techniques is to acknowledge that current KA techniques are not working well, and understand why this is the case; we hope that this paper constitutes a useful step in this direction.

## Acknowledgements

Numerous people have given us valuable comments over the past five years as we struggled with KA for NLG, too many to acknowledge here. But we would like to thank Sandra Williams for reading several drafts of this paper and considering it in the light of her own experiences, and to thank the anonymous reviewers for their very helpful comments. We would also like to thank the experts we worked with in STOP and SUMTIME-MOUSAM, without whom this work would not be possible. This work was supported by the UK Engineering and Physical Sciences Research Council (EPSRC), under grants GR/L48812 and GR/M76881, and by the Scottish Office Department of Health under grant K/OPR/2/2/D318.